%% file: main.tex
\documentclass[sigconf]{acmart}
\AtBeginDocument{%
  }

\copyrightyear{2026}
\acmYear{2026}
\setcopyright{cc}
\setcctype{by}
\acmConference[SIGIR '26]{Proceedings of the 49th International ACM SIGIR Conference on Research and Development in Information Retrieval}{July 20--24, 2026}{Melbourne, VIC, Australia}
\acmBooktitle{Proceedings of the 49th International ACM SIGIR Conference on Research and Development in Information Retrieval (SIGIR '26), July 20--24, 2026, Melbourne, VIC, Australia}
\acmDOI{10.1145/3805712.3808617}
\acmISBN{979-8-4007-2599-9/2026/07}
\usepackage{microtype}
\usepackage{graphicx}
\usepackage{subfigure}
\usepackage{booktabs} 
\usepackage{grffile} 
\usepackage{float}

\usepackage{hyperref}

\usepackage{booktabs}
\usepackage{multirow}

\usepackage{makecell}
\usepackage{tabularx}

\usepackage{amsmath}
\usepackage{mathtools}
\usepackage{amsthm}
\usepackage{enumitem}
\usepackage[most]{tcolorbox}
\usepackage[normalem]{ulem}
\useunder{\uline}{\ul}{}
\usepackage{multirow}
\usepackage{longtable}
\usepackage{makecell}
\usepackage{rotating}
\usepackage{balance} 

\newcommand{\benchmark}{TReB}

\begin{document}

\title{\benchmark{}: A Comprehensive Benchmark for Evaluating Table Reasoning Capabilities of Large Language Models}



\author{Ce Li}
\affiliation{%
  \institution{JIUTIAN Research}
  \city{Beijing}
  \country{China}
}
\email{lice@cmjt.chinamobile.com}

\author{Xiaofan Liu}
\affiliation{%
  \institution{JIUTIAN Research}
  \city{Beijing}
  \country{China}
}
\email{liuxiaofan@cmjt.chinamobile.com}

\author{Zhiyan Song}
\affiliation{%
  \institution{JIUTIAN Research}
  \city{Beijing}
  \country{China}
}
\email{songzhiyan@cmjt.chinamobile.com}

\author{Ce Chi}
\affiliation{%
  \institution{JIUTIAN Research}
  \city{Beijing}
  \country{China}
}
\email{chice@cmjt.chinamobile.com}

\author{Boshen Shi}
\affiliation{%
  \institution{JIUTIAN Research}
  \city{Beijing}
  \country{China}
}
\email{shiboshen@cmjt.chinamobile.com}

\author{Chen Zhao}
\affiliation{%
  \institution{JIUTIAN Research}
  \city{Beijing}
  \country{China}
}
\email{zhaochen@cmjt.chinamobile.com}

\author{Guanguang Chang}
\affiliation{%
  \institution{JIUTIAN Research}
  \city{Beijing}
  \country{China}
}
\email{changguanguang@cmjt.chinamobile.com}

\author{Zhendong Wang}
\affiliation{%
  \institution{JIUTIAN Research}
  \city{Beijing}
  \country{China}
}
\email{wangzhendong@cmjt.chinamobile.com}

\author{Kexin Yang}
\affiliation{%
  \institution{JIUTIAN Research}
  \city{Beijing}
  \country{China}
}
\email{yangkexin@cmjt.chinamobile.com}

\author{Xing Wang}
\authornote{Corresponding Author}
\affiliation{%
  \institution{JIUTIAN Research}
  \city{Beijing}
  \country{China}
}
\email{wangxing@cmjt.chinamobile.com}

\author{Chao Deng}
\affiliation{%
  \institution{JIUTIAN Research}
  \city{Beijing}
  \country{China}
}
\email{dengchao@cmjt.chinamobile.com}

\author{Junlan Feng}
\authornotemark[1]
\affiliation{%
  \institution{JIUTIAN Research}
  \city{Beijing}
  \country{China}
}
\email{fengjunlan@cmjt.chinamobile.com}








\renewcommand{\shortauthors}{Ce Li et al.}

\input{section/0_abstract}

\ccsdesc[300]{Information systems~Data mining}


\keywords{Table Reasoning; Benchmark; LLM Evaluation}


\maketitle

\input{section/1_introduction}

\input{section/2_dataset}

\input{section/3_eval_framework}

\input{section/4_experiment}
\input{section/6_limitations_and_future_works}
\input{section/7_conclusion}


\bibliographystyle{ACM-Reference-Format}
\balance
\bibliography{main}










\end{document}

%% file: section/0_abstract.tex
\begin{abstract}
The majority of data in businesses and industries is stored in tables, databases, and data warehouses. Reasoning with table-structured data poses significant challenges for large language models (LLMs) due to its hidden semantics, inherent complexity, and structured nature. One of these challenges is lacking an effective evaluation benchmark fairly reflecting the performances of LLMs on broad table reasoning abilities. In this paper, we fill in this gap by presenting a comprehensive table reasoning benchmark, \benchmark{}. Firstly, we propose a taxonomy to systematically measure both shallow table understanding abilities and deep table reasoning abilities, covering a total of 26 sub-tasks. We then construct a high quality dataset through a dedicated data processing and synthesis procedure. Based on these well-constructed samples, we design an evaluation framework to robustly measure table reasoning capabilities with three distinct inference modes. Experimental results with our data and framework reveal that existing LLMs still have significant room for improvement in addressing the complex and real world table related tasks. Both the dataset and evaluation framework are publicly available, with the dataset hosted on \href{https://huggingface.co/datasets/JT-LM/JIUTIAN-TReB} {\path{huggingface.co/datasets/JT-LM/JIUTIAN-TReB}}, and the framework on \href{https://github.com/JT-LM/jiutian-treb}{\path{github.com/JT-LM/jiutian-treb}}.

\end{abstract}

%% file: section/1_introduction.tex
\section{Introduction}
Table reasoning refers to the core capability of a model to interpret, manipulate, and deduce insights from tabular data through logical operations~\cite{table_llm_survey}. It is prominent in the field of natural language processing~\cite{lu2024large}, and has huge potentials in real-world applications such as Business Intelligence and Healthcare~\cite{table_mining_survey}. Traditional approaches mostly focus on encoding the semantics of tables through structure embeddings and attention mechanisms, enabling pretrained models to better understand the content of tabular data~\cite{kim2025table, su2024tablegpt2,zhu2023xtab}. In recent years, the advent of large language models (LLMs), like GPT-3.5 and GPT-4~\citep{gpt3,openai2023gpt4}, has redefined the paradigm of table reasoning methodology. Instead of relying solely on table semantic embeddings, LLMs leverage prompt engineering, external tools such as SQL and Python~\citep{wang2023mac,chai2024mceval}, and complex reasoning techniques such as chain-of-thought (CoT)~\citep{cot} to understand and analyze the tabular data. These developments have demonstrated the remarkable reasoning capability of LLMs to perform table-related data analysis, even without task-specific modifications.


Owing to the growing potential of LLMs in analyzing tabular data, several benchmarks, such as TableBench~\cite{wu2025tablebench} and RealTableBench~\cite{su2024tablegpt2}, have been developed to evaluate their reasoning capabilities. These benchmarks evaluate LLMs across multiple dimensions, including information retrieval, structural understanding, and numerical computation. Despite recent advancements, comprehensively evaluating the table-reasoning capabilities of LLMs is still challenging due to three critical reasons:

First, existing benchmarks lack a systematic taxonomy that integrates the complete workflow of a human data analyst. While prior works incorporate some basic task categorization, they fall short in modeling professional analysis skills — spanning from foundational structural understanding and precise table operations to advanced complex multi-step analysis. This deficiency limits both comprehensive capability assessment and clear error attribution.

Second, existing datasets often lack sufficient quality and real-world complexity. We assess dataset quality via label accuracy, structural integrity, and semantic consistency. Existing benchmarks contain 15\%-30\% noisy samples due to insufficient cleaning. For real-world complexity, we measure via table scale, reasoning steps, and domain authenticity. Existing data uses small tables (<50 rows, <5 columns) with single-step reasoning, failing to match enterprise-level tabular data analysis scenarios. Consequently, these datasets fail to challenge models with the deep, multi-step analysis typical of real-world applications.

Third, existing evaluation frameworks mainly rely on textual inference, where tables are provided in Markdown and models produce answers in free-form text. However, traditional textual inference can not match the rigorous demands of tabular data analysis, leading to frequent hallucinations. It is inherently error-prone for structured data analysis requiring heavy computation and mathematical modeling. In practice, it is typically instantiated as textual Chain-of-Thought (CoT)~\cite{cot}, while executable paradigms—such as Program-of-Thought (PoT)~\cite{chenprogram} and tool-integrated iterative reasoning—remain underexplored, even though they can improve precision by grounding results in code execution rather than probabilistic text generation. Consequently, existing evaluation frameworks cannot fully assess an LLM’s ability to function as a reliable, tool-using agent in authentic analytical workflows.

To address these challenges, we introduce \textbf{\benchmark{}}, a comprehensive benchmark designed to standardize the assessment of table reasoning. Our core contributions are threefold:

\begin{enumerate} 
\item \textbf{Hierarchical Taxonomy for Systematic Assessment:} We propose a fine-grained taxonomy that structures table reasoning into 6 core skills and 26 subtasks, mapping the complete analytical workflow from fundamental \textit{Natural Language Understanding} to tabular \textit{Advanced Data Analysis}. Unlike monolithic evaluations, this hierarchical design mirrors the workflow of professional data analysts, enabling precise error attribution and diagnostic insights into an LLM's specific capabilities.

\item \textbf{Full-Spectrum Data Construction:} 
To ensure comprehensive coverage of our taxonomy, we construct a high-quality dataset of 7,790 instances through a hybrid strategy. This includes the rigorous cleaning of multi-source open data (filtering for structure and semantics) and the targeted generation of synthetic data via Rule-based and LLM-based pipelines to fill coverage gaps. This approach guarantees diversity in both task complexity and reasoning paradigms.

\item \textbf{Scalable \& Robust Evaluation Framework:} 
We establish a configuration-driven framework that streamlines large-scale assessment via batch inference. Crucially, it integrates diverse inference paradigms to match task requirements: Textual Chain-of-Thought (TCoT), Program-of-Thought (PoT), and Interleaved CoT (ICoT) — a ReAct-style ~\cite{yao2023react} approach that dynamically alternates between textual analysis and code execution to solve complex, multi-step problems. For judging predictions, it employs a robust LLM-as-a-Judge mechanism, which utilizes a granular 0--10 scale to objectively evaluate the correctness of answers, overcoming the limitations of rigid string-matching metrics.
\end{enumerate}

In summary, this work systematizes table reasoning by proposing a fine-grained capability taxonomy. Based on this foundation, we introduce a holistic open-source suite comprising a high-quality dataset\footnote{\url{https://huggingface.co/datasets/JT-LM/JIUTIAN-TReB}} and an extensible evaluation framework\footnote{\url{https://github.com/JT-LM/jiutian-treb}}. Furthermore, to facilitate fair competition, we maintain a blind test leaderboard\footnote{\url{https://jt-lm.github.io/jiutian-treb}} where ground truth is withheld. These contributions aim to standardize benchmarks and accelerate the development of generalizable table reasoning models.


%% file: section/2_dataset.tex
\section{Dataset}
In this section, we first propose a hierarchical taxonomy of table reasoning capabilities. Building upon the above taxonomy, we describe our data construction pipeline, which integrates (i) curated high-quality instances obtained via systematic filtering and cleansing of open-source datasets, and (ii) synthesized examples designed to improve both the quantity and quality of the data.

\subsection{Table Reasoning Taxonomy}
\begin{table*}[htbp]
  \centering
  \caption{Taxonomy}
  \tiny
  \renewcommand{\arraystretch}{0.9} 
  \resizebox{0.85\textwidth}{!}{
    \begin{tabular}{llll} 
    \toprule
    Core Skill & Subtask & Task Description & Number of Data \\ 
    \midrule
    \multirow{6}{*}{NLU} & 
          Understanding & Evaluates LLMs' semantic comprehension capabilities & 500 \\
          & Instruction Following & Assesses LLMs' ability to follow instructions & 90 \\
          & Hallucination Evaluation & Measures LLMs' tendency to generate false information & 500  \\
          & Robustness Evaluation & Tests LLMs' stability under varied inputs & 500  \\
          & Code Generation & Evaluates LLMs' ability to generate functional code & 500 \\
          & Mathematical Reasoning & Assesses numerical reasoning capabilities & 500 \\
    \midrule
    \multirow{6}{*}{TU} & 
          Table Retrieval & Tests information retrieval from tabular data with/without prompts & 500 \\
          & Table Summary & Evaluates generation of descriptive text from tables & 500 \\
          & Table Column\_Naming & Assesses ability to infer column names from data & 500 \\
          & Table Title Naming & Evaluates generation of concise table titles & 500 \\
          & Table Fact Checking & Tests table comprehension and logical reasoning & 500 \\
          & Table Plausibility Verification & Assesses table content validity using prior knowledge & 15\\
    \midrule
    \multirow{2}{*}{TBO} & 
          Table Query & Evaluates precise and fuzzy query capabilities on tabular data & 500 \\
          & Table Selection & Tests table reasoning filtering (exact and semantic-based) & 500 \\
    \midrule
    \multirow{2}{*}{TCO} & 
          Table General Operations & Assesses basic statistical computations on tables & 500 \\
          & Table Domain-Specific Operations & Evaluates domain-specific formula applications & 239\\
    \midrule
    \multirow{4}{*}{DA} & 
          Table Outlier Detection & Tests identification of anomalous data points & 43 \\
          & Table Correlation Analysis & Evaluates inter-column relationship analysis & 63 \\
          & Table Hypothesis Testing & Assesses statistical testing capabilities & 42 \\
          & Table Distribution Testing & Evaluates probability distribution analysis & 500 \\
    \midrule
    \multirow{6}{*}{ADA} & 
          Multi-step Retrieval & Tests multi-step computation and information retrieval & 49 \\
          & Multi-step Fact Checking & Evaluates multi-step fact verification & 61 \\
          & Multi-step Operations & Assesses complex table-based calculations & 61 \\
          & Multi-step Correlation Analysis & Tests advanced correlation analysis & 49 \\
          & Multi-step Hypothesis Testing & Evaluates complex hypothesis testing & 61 \\
          & Multi-step Conditional Calculation & Assesses conditional computations based on derived tables & 17 \\
    \bottomrule
    \end{tabular}%
    }
  \label{eval-metric-frame}%
\end{table*}%

Existing evaluations for table reasoning predominantly rely on single-task datasets or limited multi-task benchmarks, failing to capture the full breadth of LLM capabilities~\cite{table_mining_survey,lu2024large,table_llm_survey}. Furthermore, real-world tabular tasks require a progression from atomic operations to composite logic. By structuring the evaluation from \textit{Basic Understanding} to \textit{Operational Execution} and finally to \textit{Insight Generation}, our taxonomy mirrors the human cognitive workflow of data analysis. Motivated by this process, we propose a fine-grained hierarchical taxonomy that structures table reasoning into a comprehensive capability spectrum. As detailed in Table \ref{eval-metric-frame}, this taxonomy ranges from fundamental language understanding to advanced data analysis, encompassing 6 core skills and 26 distinct subtasks.

\begin{itemize}
    \item Natural Language Understanding (NLU):
    Evaluates foundational language skills required for tabular reasoning, including instruction following, robustness to perturbations, coding, math, etc.
    \item Table Understanding (TU): 
    Assesses structural grasp of tabular data, focusing on schema inference (e.g., column naming), content summarization, and factual verification of table entries.
    \item Table-Based Operations (TBO):
    Tests the ability to perform precise information retrieval and filtering (row/column selection) based on both exact and fuzzy query constraints.
    \item Table Calculation Operations (TCO):
    Examines numerical reasoning skills, ranging from fundamental statistical computations to the application of domain-specific formulas.
    \item Data Analysis (DA):
    Focuses on deriving analytical insights from data distributions, including outlier detection, correlation analysis, and statistical hypothesis testing.
    \item Advanced Data Analysis (ADA):
    Challenges the model with complex, multi-step reasoning tasks that require logical planning to execute composite operations (e.g., multi-step retrieval and conditional calculations).
\end{itemize}

In summary, this hierarchical taxonomy moves beyond flat metrics to disentangle atomic capabilities. It transforms evaluation from a black-box assessment into a structured inquiry.

\subsection{Open-source Data Collection}

To rigorously evaluate models under our taxonomy, we first collect public datasets from multiple sources. This ensures the coverage across most of the capability spectrum, from atomic semantic understanding to tabular reasoning. In particular, we design a construction pipeline including capability-aligned acquisition followed by a rigorous cleaning protocol.

Firstly, to assess the fundamental natural language understanding layer of the taxonomy, we curated 59,901 instances from ten representative datasets~\cite{wang2024mmlu,sakaguchi2021winogrande,hendrycks2021measuring,cobbe2021training,liu2024mathbench,zhuo2024bigcodebench,lai2023ds,liang2023uhgeval,jing2023followeval,wang2021adversarial}. Post-processing included deduplication, format normalization, and English-Chinese bilingual translation to ensure linguistic diversity.

Then, to further support the tabular reasoning, we harvested over 2 million raw samples via a hybrid strategy of keyword retrieval and manual curation spanning two decades of literature (e.g., Web of Science, Google Scholar). We targeted keywords such as ``table structure recognition'', ``cell-level QA'', and ``table fact checking''. After filtering for representativeness and format compatibility (Markdown/HTML/CSV), we integrated 29 canonical benchmarks (e.g., AIT-QA~\cite{aitqa}, ToTTo~\cite{totto}, HybridQA~\cite{hybridqa}, and TableBench~\cite{wu2025tablebench}) sourced from Wikipedia, financial reports, and academic publications.

To guarantee evaluation reliability, we implemented a rigorous multi-stage cleaning pipeline addressing both structural and semantic dimensions. On the structural level, we enforce strict constraints by capping the context at 32K characters, consistent with the context window limit of most open-source LLMs. We remove empty or overly sparse tables, and simplify nested headers to improve efficiency and readability. On the semantic level, we deployed a \textbf{Consensus-Based Quality Control Framework} where a judge model evaluates inferences from three candidate LLMs against ground truth, supplemented by a human-in-the-loop review for ambiguous cases. This rigorous filtration process distinguished high-quality data from noise, ultimately yielding \textbf{0.75 million verified instances} that serve as a robust foundation for systematic table reasoning assessment. 

Finally, to map each sample to the defined taxonomy, we clearly define the evaluation criteria and task boundaries, and leverage LLMs to automatically classify such samples into the corresponding subtasks, ensuring precise alignment and avoiding overlaps across subtasks.

\subsection{Data Synthesis}

To bridge the coverage gap where open-source datasets fail to encompass the full spectrum of our 26-subtask taxonomy, we implemented a hybrid data synthesis pipeline.

First, for 10 structured sub-tasks requiring consistent formats (e.g., Table Retrieval, Query, Selection), we employ \textit{Rule-based generation}. This involves selecting numeric tables with computable properties and applying logic-driven templates to synthesize QA pairs for deterministic operations such as filtering, sorting, and statistical computation. 

Then, for 4 sub-tasks demanding semantic flexibility (e.g., \textit{Table Summary, Plausibility, Domain-specific Operations}), we utilize \textit{LLM-based generation}. This strategy operates in two modes: generating QA pairs from existing tables, or co-synthesizing table-QA pairs when necessary. All LLM-generated outputs undergo a rigorous dual-discriminator verification process; only samples achieving unanimous consensus on accuracy, relevance, and coverage are retained.

\begin{figure*}[htbp]
\centerline{\includegraphics[width=0.90\textwidth]{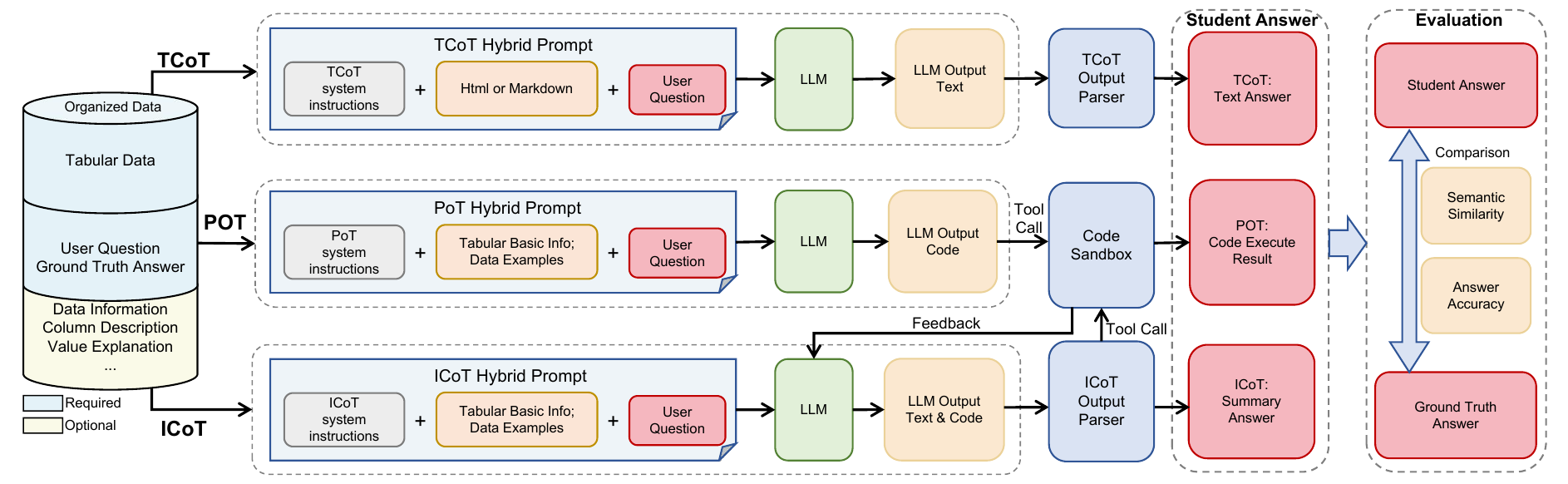}}
\caption{Evaluation Framework Overview.}
\label{fig:framework}
\end{figure*}

In particular, to simulate the sequential complexity of real-world data analysis, we developed a pipeline that synthesizes multi-turn QA pairs requiring chained reasoning. Covering 6 advanced subtasks (e.g., Multi-step Retrieval, Conditional Calculation), this method constructs 'harder' samples that compel models to perform iterative verification and calculation, mirroring the workflow of a human expert. Such pipeline includes four parts: 1) \textbf{Collect complex tables}. We define a 'complex table' as one that exhibits high dimensionality ($>100$ rows, $>10$ columns) and analytical richness. Specifically, eligible tables must possess semantically clear headers, at least two categorical columns for filtering, and four numeric columns with non-zero variance for computation. Furthermore, we enforce strict quality controls, excluding tables with $>80\%$ missing values or significant textual noise. 2) \textbf{Model complex reasoning logic}. We model analytical reasoning as a Directed Acyclic Graph (DAG), where nodes represent atomic operators and edges define data dependencies. To structure the workflow, we categorize these operators into three hierarchical tiers: \textit{Selection \& Filtering} (defining analytical scope), \textit{Aggregation \& Transformation} (synthesizing metrics), and \textit{Logical Derivation} (computing final insights). Consequently, a specific analyst workflow is formalized as a random walk over this operator graph. 3) \textbf{Generate multi-turn reverse questions}. Based on the sampled workflow, we prompt two LLMs in an adversarial manner: one for question generation and another as a discriminator, which evaluates semantic plausibility, logical consistency, and answerability. For questions that pass the discriminator’s criteria, we further refine the questions from the perspective of real users to enhance their authenticity. 4) \textbf{Generate answers}. For each table and its associated question pair, multiple candidate answers are generated using LLM. These candidates then undergo human review to ensure the accuracy and reliability of the final accepted answers.

Finally, we sample the verified high-quality data and release a taxonomy-aligned subset (as shown in Table \ref{eval-metric-frame}) as the benchmark for evaluating large language models. We have fully open-sourced the benchmark dataset, which comprises 7,790 instances covering 6 core skills and 26 subtasks, with an average of 64 rows and 8 columns per table, covering domains such as finance, education, healthcare, and enterprise. The complete dataset are provided in our open repository.

%% file: section/3_eval_framework.tex
\section{Evaluation Framework}

The evaluation framework is designed to systematically assess the performance of LLMs in table reasoning. In particular, the system operates on a configuration-driven basis, enabling rapid switching between models and task sets via simple config modifications. To handle the computational load, the framework employs batch inference for high throughput. Furthermore, to maximizes GPU utilization, it features a decoupled two-stage architecture: inference and judgement are executed sequentially. In the following part we detail the framework's workflow from the Inference phase and the judge Phase.

As shown in Figure \ref{fig:framework}, the framework begins with loading batch of samples from the dataset built above. Each sample is a four-element tuple containing (Table, Question, Groundtruth, Extra-Info). Then, the framework incorporates three distinct inference modes: Textual Chain-of-Thought (TCoT), Programmatic Chain-of-Thought (PoT), and Interleaved Chain-of-Thought (ICoT). For TCoT, a table is organized as a markdown or html string which inserted into user question. Then the model operates on textual reasoning, generating answers in plain text. For PoT, we inject the basic schema information into user question and then the model produces executable code, which is further executed in the sandbox to derive the final answer. The ICoT mode enables models to perform multi-step reasoning by combining textual explanations and programmatic outputs in an iterative process, encouraging the model to engage in planning, iterative step-by-step reasoning, and self-reflection. The table storage path is provided, forcing model to read data by python code. Specifically, the textual reasoning content is wrapped within `<think></think>' tags, the python code is wrapped with `\verb|```python| \verb|```|', and the executed result is wrapped with `\verb|```CodeExecuteResult| \verb|```|'.

Following the inference phase, the framework evaluates model outputs using an \textit{LLM-as-a-Judge} paradigm~\cite{gu2024survey,zheng2023judging,dubois2023alpacafarm, DBLP:conf/acl/ChiangL23}. 
We adopt this approach to overcome the limitations of rigid string-matching metrics (e.g., Exact Match, ROUGE), which often fail to capture the semantic validity of open-ended table reasoning tasks. These traditional metrics risk penalizing stylistically distinct but factually correct responses. 
To address this, the judger acts as a robust surrogate for human reviewers, assigning an integer score on a scale of 0 to 10. A score of 0 denotes a completely incorrect answer, while 10 indicates a perfect match with the ground truth. Crucially, this granular scale accommodates \textit{partial correctness}: for queries requiring multiple answers, models that retrieve incomplete but valid subsets are awarded intermediate scores, ensuring a more objective assessment of reasoning capabilities.

%% file: section/4_experiment.tex
\section{Experiments}
In this section, we conduct a comprehensive evaluation of over 20 state-of-the-art LLMs on our benchmark, providing an in-depth analysis of their performance across various table reasoning tasks.

\begin{table*}[htbp]
  \centering
    \caption{Overall Experimental Results with LLM-as-a-judge}
    \label{Overall Experimental Results with LLM-as-a-judge}
    \renewcommand{\arraystretch}{0.9} 
      \resizebox{0.9\textwidth}{!}{
        \begin{tabular}{lccccccccccccccccr}
            \toprule
            \multicolumn{1}{c}{\multirow{2}[4]{*}{\textbf{Model Name}}} & \multicolumn{1}{c}{\textbf{NLU}} & \multicolumn{3}{c}{\textbf{TU}} & \multicolumn{3}{c}{\textbf{TBO}} & \multicolumn{3}{c}{\textbf{TCO}} & \multicolumn{3}{c}{\textbf{DA}} & \multicolumn{3}{c}{\textbf{ADA}} & \multicolumn{1}{c}{\multirow{2}[4]{*}{\textbf{Overall}}} \\
        \cmidrule(lr){2-2} \cmidrule(lr){3-5} \cmidrule(lr){6-8} \cmidrule(lr){9-11} \cmidrule(lr){12-14} \cmidrule(lr){15-17}          & \multicolumn{1}{c}{\textbf{TCoT}} & \multicolumn{1}{c}{\textbf{TCoT}} & \multicolumn{1}{c}{\textbf{PoT}} & \multicolumn{1}{c}{\textbf{ICoT}} & \multicolumn{1}{c}{\textbf{TCoT}} & \multicolumn{1}{c}{\textbf{PoT}} & \multicolumn{1}{c}{\textbf{ICoT}} & \multicolumn{1}{c}{\textbf{TCoT}} & \multicolumn{1}{c}{\textbf{PoT}} & \multicolumn{1}{c}{\textbf{ICoT}} & \multicolumn{1}{c}{\textbf{TCoT}} & \multicolumn{1}{c}{\textbf{PoT}} & \multicolumn{1}{c}{\textbf{ICoT}} & \multicolumn{1}{c}{\textbf{TCoT}} & \multicolumn{1}{c}{\textbf{PoT}} & \multicolumn{1}{c}{\textbf{ICoT}} &  \\
            \midrule
            \multicolumn{18}{c}{\textit{\textbf{General LLMs}}} \\
            \midrule
            Llama-3.1-8B-Instruct~\cite{dubey2024llama} & 61.46  & 49.53  & 45.04  & 47.06  & 39.22  & 50.11  & 55.76  & 41.51  & 55.12  & 50.20  & 39.68  & 51.97  & 50.89  & 31.36  & 24.01  & 13.62  & 44.16  \\
            Llama-3.1-70B-Instruct~\cite{dubey2024llama} & 70.75  & 64.38  & 66.47  & 57.93  & 57.74  & 72.20  & 48.04  & 65.75  & 77.69  & 61.46  & 48.58  & 61.80  & 51.91  & 40.29  & 42.48  & 27.50  & 57.18  \\
            Qwen2.5-7B-Instruct~\cite{qwen2.5} & 64.66  & 58.79  & 57.98  & 68.26  & 41.46  & 56.99  & 61.23  & 51.61  & 62.71  & 73.63  & 44.88  & 52.17  & 61.84  & 37.13  & 28.08  & 44.63  & 54.13  \\
            Qwen2.5-72B-Instruct~\cite{qwen2.5} & 75.87  & 72.39  & 64.46  & 87.07  & 66.56  & 71.05  & 72.06  & 73.18  & 80.93  & 85.25  & 54.05  & 59.69  & 70.97  & 42.14  & 46.49  & 61.25  & 67.71  \\
            Mistral-7B-Instruct-v0.3~\cite{jiang2023mistral} & 47.47  & 40.22  & 37.33  & 25.14  & 32.57  & 50.09  & 37.18  & 35.28  & 45.46  & 35.61  & 37.20  & 23.96  & 27.09  & 18.89  & 15.14  & 16.05  & 32.79  \\
            \midrule
            \multicolumn{18}{c}{\textit{\textbf{Code Optimized LLMs}}} \\
            \midrule
            Qwen2.5-Coder-7B-Instruct~\cite{hui2024qwen2} & 61.87  & 55.51  & 57.92  & 64.30  & 44.36  & 61.03  & 66.86  & 47.59  & 65.17  & 69.26  & 39.99  & 57.73  & 62.99  & 35.99  & 33.97  & 45.50  & 54.38  \\
            Deepseek-Coder-7B-Instruct-v1.5~\cite{guo2024deepseekcoder} & 16.63  & 21.82  & 19.36  & 12.38  & 26.83  & 36.69  & 24.33  & 19.93  & 30.11  & 19.26  & 14.65  & 13.46  & 14.28  & 13.27  & 3.62  & 13.77  & 18.77  \\
            Deepseek-Coder-33B-Instruct~\cite{guo2024deepseekcoder} & 26.59  & 35.09  & 49.06  & 24.39  & 39.40  & 57.01  & 54.72  & 29.16  & 42.74  & 30.91  & 24.77  & 53.87  & 33.54  & 10.94  & 26.96  & 19.60  & 34.92  \\
            Seed-Coder-8B-Instruct~\cite{seed2025seedcoderletcodemodel} & 45.18  & 50.65  & 57.21  & 59.61  & 44.06  & 66.58  & 66.33  & 42.66  & 67.57  & 68.67  & 38.68  & 62.94  & 65.62  & 32.59  & 38.05  & 42.27  & 53.04  \\
            Yi-Coder-9B-Chat~\cite{ai2024yi} & 43.64  & 32.76  & 50.61  & 39.70  & 30.59  & 56.24  & 51.56  & 32.49  & 58.34  & 54.58  & 23.07  & 48.17  & 51.50  & 21.41  & 28.72  & 20.62  & 40.25  \\
            \midrule
            \multicolumn{18}{c}{\textit{\textbf{Deep Thinking LLMs}}} \\
            \midrule
            Deepseek-R1-Distill-Qwen-7B~\cite{deepseekai2025deepseekr1incentivizingreasoningcapability} & 51.28  & 49.28  & 33.86  & 54.58  & 56.33  & 48.24  & 52.49  & 62.01  & 43.28  & 56.50  & 49.19  & 41.19  & 57.34  & 22.86  & 18.02  & 36.89  & 45.83  \\
            Deepseek-R1-Distill-Qwen-14B~\cite{deepseekai2025deepseekr1incentivizingreasoningcapability} & 61.78  & 62.29  & 66.71  & 81.18  & 73.07  & 65.02  & 68.66  & 67.10  & 51.27  & 79.11  & 53.87  & 46.51  & 61.98  & 34.81  & 31.34  & 51.26  & 59.75  \\
            Deepseek-R1-Distill-Qwen-32B~\cite{deepseekai2025deepseekr1incentivizingreasoningcapability} & 63.04  & 69.19  & 65.33  & 86.18  & 78.23  & 71.92  & 70.51  & 68.34  & 62.82  & 69.76  & 56.36  & 67.90  & 61.33  & 32.14  & 43.22  & 52.51  & 63.67  \\
            Deepseek-R1-Distill-Llama-8B~\cite{deepseekai2025deepseekr1incentivizingreasoningcapability} & 53.17  & 56.13  & 34.86  & 54.48  & 55.23  & 39.07  & 49.60  & 57.29  & 36.86  & 52.17  & 49.24  & 13.14  & 40.90  & 20.32  & 6.40  & 25.19  & 40.25  \\
            Deepseek-R1-Distill-Llama-70B~\cite{deepseekai2025deepseekr1incentivizingreasoningcapability} & 65.21  & 67.88  & 67.51  & 86.98  & 74.65  & 69.16  & 69.23  & 80.07  & 76.14  & 86.08  & 59.35  & 66.64  & 73.03  & 37.27  & 44.68  & 54.04  & 67.37  \\
            QwQ-32B~\cite{qwq32b} & 71.39  & 71.51  & 72.62  & 91.02  & 80.57  & 78.23  & 74.00  & 74.47  & 73.31  & 75.36  & 51.79  & 72.54  & 60.68  & 45.06  & 54.79  & 64.88  & 69.51  \\
            Qwen3-8B~\cite{qwen3technicalreport} & 62.60  & 67.27  & 70.68  & 82.93  & 80.87  & 69.06  & 70.51  & 71.25  & 66.31  & 71.58  & 58.16  & 55.51  & 61.14  & 52.76  & 44.34  & 57.41  & 65.15  \\
            Qwen3-14B~\cite{qwen3technicalreport} & 68.07  & 68.80  & 64.08  & 87.33  & 79.26  & 76.16  & 77.51  & 73.03  & 69.27  & 73.63  & 58.00  & 66.49  & 66.18  & 48.28  & 50.55  & 56.90  & 67.72  \\
            Qwen3-32B~\cite{qwen3technicalreport} & 67.72  & 67.84  & 71.16  & 90.89  & 81.19  & 75.83  & 77.56  & 73.32  & 70.91  & 75.07  & 62.23  & 68.63  & 67.55  & 49.63  & 57.34  & 62.70  & 69.97  \\
            \midrule
            \multicolumn{18}{c}{\textit{\textbf{Math Optimized LLMs}}} \\
            \midrule
            Kimina-Prover-Preview-Distill-7B~\cite{kimina_prover_2025} & 22.69  & 11.53  & 0.15  & 9.85  & 19.50  & 0.40  & 16.58  & 24.94  & 0.70  & 16.31  & 12.67  & 0.05  & 11.43  & 8.04  & 0.00  & 8.66  & 10.22  \\
            Qwen2.5-Math-7B-Instruct~\cite{yang2024qwen25mathtechnicalreportmathematical} & 40.17  & 22.35  & 13.48  & 41.38  & 34.68  & 26.22  & 38.29  & 37.61  & 31.01  & 51.51  & 18.68  & 20.04  & 39.08  & 12.09  & 2.25  & 19.57  & 28.03  \\
            Qwen2.5-Math-72B-Instruct~\cite{yang2024qwen25mathtechnicalreportmathematical} & 56.98  & 53.56  & 46.57  & 76.98  & 58.35  & 48.73  & 59.40  & 61.20  & 58.62  & 79.27  & 35.23  & 35.51  & 58.50  & 17.74  & 23.70  & 39.31  & 50.60  \\
            Deepseek-Math-7B-Instruct~\cite{shao2024deepseekmath} & 28.85  & 21.02  & 7.77  & 28.98  & 15.91  & 4.25  & 8.41  & 20.61  & 10.91  & 11.45  & 20.34  & 9.50  & 18.02  & 12.63  & 1.30  & 11.02  & 14.43  \\
            \midrule
            \multicolumn{18}{c}{\textit{\textbf{Table Reasoning Optimized LLMs}}} \\
            \midrule
            TableGPT2-7B~\cite{su2024tablegpt2} & 60.83  & 58.97  & 64.61  & 73.82  & 48.38  & 59.79  & 65.43  & 57.05  & 71.94  & 75.58  & 44.14  & 51.24  & 66.80  & 32.06  & 33.21  & 50.68  & 57.16  \\
            Table-R1-SFT-7B~\cite{yang2025table} & 71.58  & 62.04  & 53.52  & 25.10  & 68.21  & 54.49  & 15.70  & 71.25  & 59.87  & 13.57  & 37.08  & 41.26  & 33.75  & 35.64  & 25.95  & 19.95  & 43.06  \\
            Table-R1-Zero-7B~\cite{yang2025table} & 66.18  & 64.36  & 48.99  & 77.99  & 54.56  & 36.01  & 58.57  & 62.06  & 50.28  & 76.16  & 50.32  & 48.95  & 63.79  & 28.21  & 32.32  & 45.36  & 54.01  \\
            \bottomrule
            \end{tabular}%
      }
\end{table*}
\begin{figure*}[htpb]
\centerline{\includegraphics[width=0.9\textwidth]{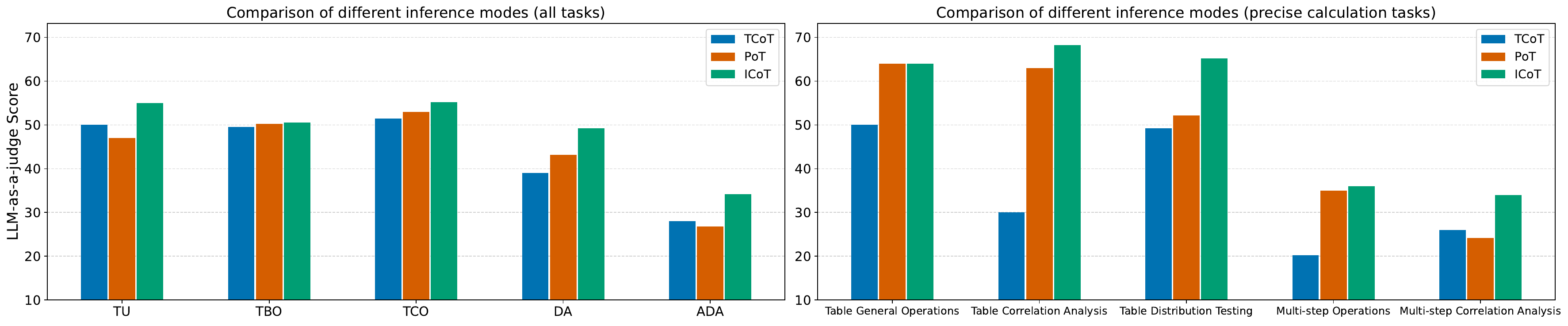}}
\caption{Comparison of table reasoning capability under different reasoning modes.}
\label{fig:compare reasonoing mode}
\end{figure*}

\subsection{Experiment Setup}
\subsubsection{LLMs}
\label{LLMs}
We evaluate a total of 26 LLMs, covering a diverse range of models designed for different purposes. These include general LLMs, code-optimized LLMs, deep thinking LLMs, math and structured data analysis optimized LLMs, and LLMs specifically fine-tuned for table reasoning tasks. The evaluated models range from 7B to 72B parameters. All models reported in this paper are open-source, as their model sizes and architectural details are publicly available and thus facilitate fair and transparent comparisons. Evaluation results for closed-source models are provided in our GitHub repository for reference.

\subsubsection{Inference Mode}
We evaluate models using three inference modes: TCoT, PoT, and ICoT. For NLU core skills (non-tabular structural tasks), only TCoT is used as these tasks focus on language understanding rather than code-based table manipulation. In TCoT, models receive table content in Markdown/HTML formats and directly generate answers. PoT and ICoT modes do not provide the model with plaintext table content. Instead, the models have to write codes to read the table, extract the required information, and finally answer the questions.

\subsubsection{Evaluation Metrics}
In the following experiments, we primarily use LLM-as-a-judge~\cite{zheng2023judging} to evaluate model performance across tasks. LLM-as-a-judge evaluates semantic similarity and answer accuracy. For LLM-as-a-judge, we utilize the Qwen2-72B-Instruct~\cite{qwen2} model (a distinct variant from the evaluated Qwen2.5 series with independent parameter initialization and training objectives) which has been fully trained with RLHF to align with human preferences, to evaluates semantic similarity and answer accuracy. This model is excluded from the evaluated model set and uses a separate inference pipeline to minimize family-related evaluation bias. Notably, since PoT produces code as its output, we use the result of code execution as the model answer. This means that if the code fails to execute, the test sample is automatically assigned a score of zero.

\begin{figure*}[htpb]
\centerline{\includegraphics[width=0.9\textwidth]{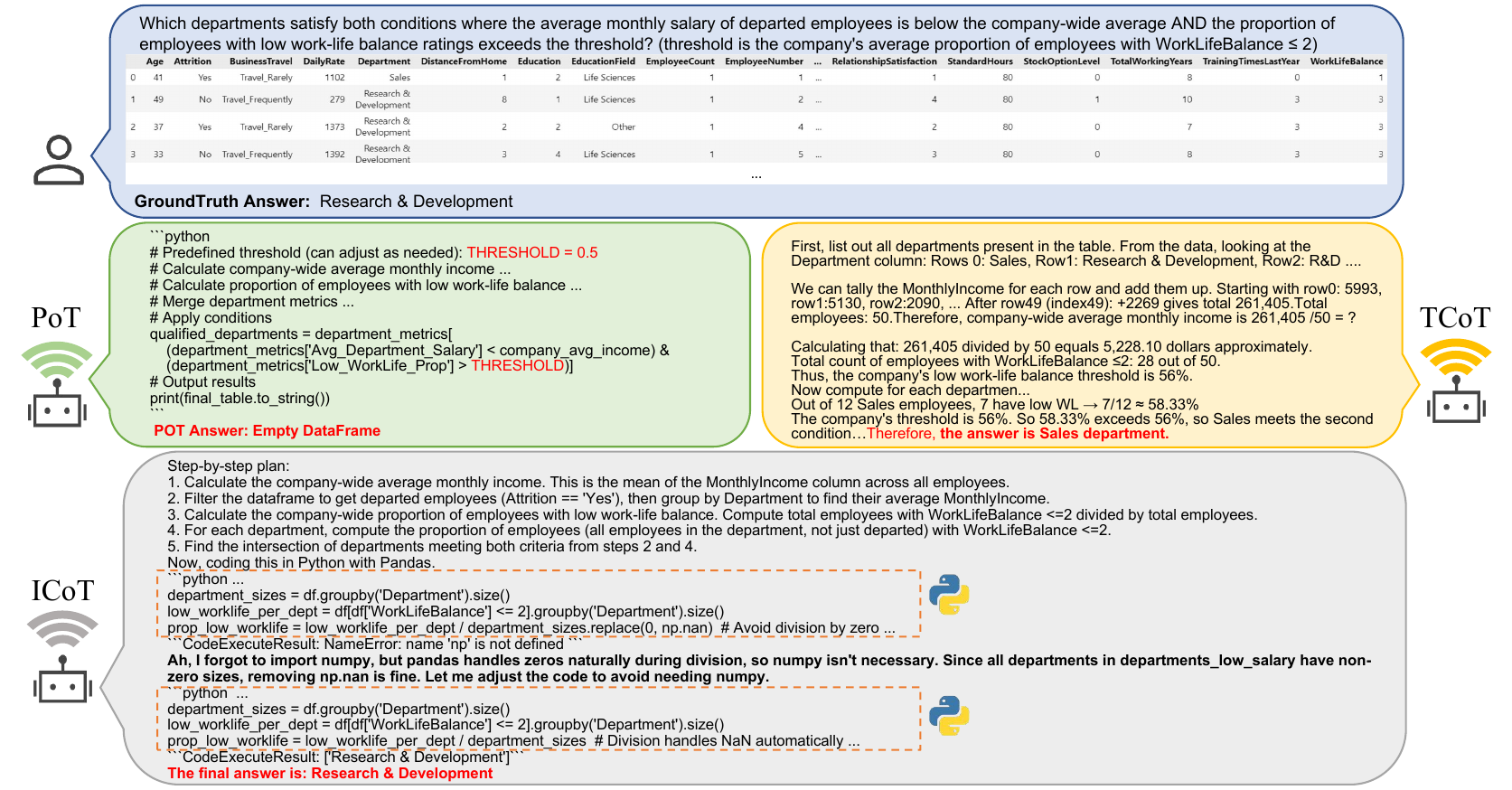}}
\caption{A representative case that demonstrates the differences among three inference modes.}
\label{fig:case2}
\end{figure*}

\subsection{Experimental Results}

\subsubsection{Overall Performance Analysis} 
We summarize the average performance of all models across all tasks, with scores normalized to a 0--100 range (as shown in Table \ref{Overall Experimental Results with LLM-as-a-judge}). We conduct human evaluation on 5\% randomly sampled instances and the correlation between human scores and LLM-as-a-judge scores is above 0.85, demonstrating consistent judgment. 

Two trends can be observed from the results: (1) Models scoring higher on Natural Language Understanding (NLU) and Table Understanding (TU) generally perform better on other tasks, as these two skills underpin question comprehension and table content retrieval. Both of them serve as foundational skills for more advanced capabilities. (2) The datasets of Advanced Data Analysis (ADA) are more challenging versions of Table Basic (TBO), Computational Operations (TCO), and Data Analysis (DA). Therefore, the scores of ADA are generally lower than other tasks due to the complexity of both the tables and the questions. Through error analysis, we observe that even the top-performing models still face significant challenges in multi-step conditional calculation. In particular, several complex tasks under ADA are universally difficult for all evaluated models. This result indicates that current LLMs still have substantial room to improve on complex, multi-step table reasoning.

We also compared the performance of the five types of LLMs. Generally, within the same model series, larger models outperform their smaller counterparts. For example, the Llama-3.1-70B model achieves better results than the Llama-3.1-8B, and the Qwen2.5-72B model outperforms the Qwen2.5-7B. These trends are also broadly consistent with the scaling laws observed for LLMs. The results also show that Deep Thinking LLMs achieve the highest overall performance. Their advanced reasoning capabilities for complex problems and self-reflection abilities enable them to consistently achieve top scores across the table reasoning tasks. By the way, in practice, performance depends on not only model type but multiple factors, including model size, training strategy, and inference configuration. Nevertheless, our evaluation still reveals the specific tabular reasoning capabilities of different types of LLMs.

\subsubsection{Performance Analysis by Inference Mode}
We analyze model performance across the three inference modes: TCoT, PoT, and ICoT. As shown in Figure \ref{fig:compare reasonoing mode}, the left panel provides an overall overview, where the results are averaged across all sub-tasks under each table reasoning skill and grouped by the three inference modes. Overall, models under the ICoT mode achieve better performance, particularly in DA and ADA tasks, outperforming the traditional TCoT approach. This demonstrates the potential of the ICoT paradigm in handling table reasoning tasks. In fact, the TCoT and ICoT modes differ fundamentally in how they handle table content: TCoT directly inputs the table in Markdown or HTML format, while ICoT enables the model to actively explore the table content through iterative interactions. This distinction becomes critical when dealing with large tables. Due to the context window limitations, TCoT struggles to process the entire table content, whereas ICoT, being independent of context size, is unaffected by table size and can dynamically query the table to retrieve relevant information. 

The right panel of Figure \ref{fig:compare reasonoing mode} focuses on performance evaluations for tasks requiring precise calculations. A notable trend emerges: TCoT underperforms in calculation-intensive tasks. This is because TCoT fundamentally relies on token-based predictions and lacks the capability to perform precise computations. In contrast, PoT and ICoT excel in such tasks by generating and execute code in a sandbox environment. Notably, the ICoT mode enables iterative code generation, allowing the model to self-reflect and correct errors. This iterative coding and execution mechanism enables ICoT to excel in handling complex numerical calculation table reasoning tasks.

\subsubsection{Case Study}\label{Case Study}
Figure \ref{fig:case2} illustrates a multi-step table reasoning example, alongside the inference processes of the QwQ-32B model across three modes. In PoT, the model generates erroneous code due to an incorrectly defined threshold and output a wrong answer.
In TCoT, the mode attempts to solve the problem through a step-by-step textual reasoning process. However, TCoT may be inappropriate for computation-intensive tasks. First, the limited token budget restricts the size of the table that can be processed, rendering TCoT ineffective for large-scale tabular data. Second, complex operations on tables tend to result in reasoning errors of arithmetic operations, further limiting the applicability of TCoT to such tasks.
Contrastly, the ICoT method demonstrates robustness through its iterative refinement capability. Although the model initially produces incorrect code, with receiving feedback of the incorrect code, which enables the model to revise its initial code, fix the error, and ultimately produce the correct answer.

The comparative analysis of this case demonstrates that the ICoT offers greater tolerance for intermediate errors and provides the model with opportunities for self-reflection and self-correction.

%% file: section/6_limitations_and_future_works.tex
\section{Limitations and Future Work}

\subsection{Limitations}

One limitation of our framework is its reliance on LLM-as-a-judge, which may inadvertently introduce biases. These biases stem from the inherent tendencies of LLMs to favor certain reasoning styles or answer formats over others. We have taken multiple steps to mitigate these biases by carefully designing and refining system prompts to ensure neutrality and consistency in scoring. Despite this, extensive experimental analyses show that in the vast majority of cases, LLM-as-a-judge achieves more objective results compared to other evaluation methods.

Furthermore, since a considerable portion of our dataset is collected from public web sources, we cannot verify whether the evaluated LLMs have been trained on these exact or highly similar data points, which may lead to unintentional data leakage and overestimated performance.

\subsection{Future Work}
Several promising directions for future work can further enhance the scope and utility of this benchmark:

\textbf{Complex Excel Tables and Multi-Table Scenarios:} Another important direction is the incorporation of more sophisticated datasets, including complex Excel tables and multi-table reasoning tasks. These additions would enable the evaluation of models' abilities to handle inter-table relationships, perform advanced operations across multiple datasets, and answer questions of higher complexity. By simulating real-world challenges, this extension would allow for a more comprehensive assessment of model capabilities in practical table mining applications.

\textbf{Enhanced Tool Integration:} Future work could also focus on extending the framework to evaluate models' ability to integrate with external tools, such as databases, APIs, or advanced computational systems. This would enable the benchmark to assess how effectively models can utilize external resources to solve highly complex or domain-specific tasks that go beyond the limits of standalone reasoning.


%% file: section/7_conclusion.tex
\section{Conclusion}
In this work, we presented a comprehensive benchmark \benchmark{} designed to standardize the evaluation of table reasoning. Guided by a fine-grained capability taxonomy, we introduced a high-quality dataset and a practical evaluation framework to support diverse inference strategies. Our extensive experiments yield critical insights into the current landscape of LLMs: (1) the Advanced Data Analysis tasks serve as a rigorous touchstone, effectively probing models' intrinsic problem-solving abilities; (2) Deep Thinking LLMs demonstrate superior overall performance by leveraging self-reflection to handle complex reasoning; and (3) ICoT-based approaches exhibit a distinct competitive edge in computation-intensive scenarios. By open-sourcing these resources and establishing a held-out leaderboard, we aim to facilitate fair comparison and accelerate the development of more robust, generalizable tabular models.